\documentclass[11pt]{article}

\usepackage{acl} %TODO: [review]

\usepackage{times}
\usepackage{latexsym}

\usepackage[T1]{fontenc}
\usepackage{amsmath}
\usepackage[utf8]{inputenc}

\usepackage{microtype}

\usepackage{inconsolata}

\usepackage{tcolorbox}
\usepackage{comment}
\tcbset{colback=white, colframe=black, boxrule=0.6pt, arc=2pt}
\tcbuselibrary{skins, breakable}

\usepackage{graphicx}
\usepackage{xcolor}
\usepackage{relsize}
\usepackage{booktabs}
\usepackage{comment}
\usepackage{makecell}
\usepackage{multirow}
\usepackage{array}

\usepackage{caption}
\usepackage{tabularx}

\newcommand{\madrs}{\texttt{MADRS} Pipeline }
\newcommand{\madrsnospace}{\texttt{MADRS} Pipeline}

\title{The \madrsnospace: Supporting Depression Assessment in Clinical Trials}
\author{\\
  {\bf \textsuperscript{*}Mila Fodor} \and
  {\bf \textsuperscript{*}Katalin Ócsai} \and 
  {\bf \textsuperscript{*+}Francesco Periti} \and
  {\bf Rien Sonck} \and \\
  {\bf Alex Boudreau} \\
  {\footnotesize\texttt{\textsuperscript{*}These authors contributed equally.} \  \textsuperscript{+}Corresponding author} \\
  \textit{Clario, part of Thermo Fisher Scientific} \\
  \texttt{name.surname@clario.com}
}

\begin{document}
\maketitle
\begin{abstract}
Depression is a major mental disorder for which diagnosis relies primarily on clinical assessments. Automated methods to support its detection via the psychiatric MADRS scale are getting more and more attention. While existing solutions primarily focus on detecting the disorder from different text sources (e.g., online text, social media), there is still limited support for clinical trials, where clinical assessments are conducted through structured interviews based on standard guidelines such as SIGMA. In this work, we develop a LLM pipeline specifically designed to support clinicians in supporting the assessment of depression in patients enrolled in clinical trials. Our pipeline converts audio interviews into transcripts, maps them into the ten MADRS symptom items, estimates their severity, and identify problematic clinical ratings associated with them. Evaluation on real clinical interviews shows a strong overall correlation of \textbf{0.867} with expert ratings, providing interpretable support for future assessments in clinical trials. 
\end{abstract}

\section{Introduction}
Depression is a major mental health disorder and a significant risk factor for suicidal behavior~\cite{weinberger2018trends}. In the past few years, notable scientific progress has been made in understanding its molecular and neurobiological underpinnings~\cite{by2025lncrna,chourpiliadis2024jama,fries2023molecular}. Despite these advances, definitive biological markers and validated diagnostic tests are still lacking~\cite{lakhan2010biomarkers}. As a result, %clinical diagnosis in 
\textit{clinical trials} for depression continues to rely on structured interviews and longitudinal assessments where clinicians are required to interpret behavioral %and emotional 
symptoms~\cite{stuart2014comparison}.

The quality of these assessments is particularly critical for trial outcomes, %clinical diagnosis, 
as it can make the difference between a failed study and one in which the target drug separates from placebo~\cite{kobak2005interview}. To improve consistency and reliability across sites and clinicians, symptom severity is quantified using validated rating scales, such as the \textit{Montgomery–Åsberg Depression Rating Scale} (\textbf{MADRS})~\cite{montgomery1979new}, and interviews are conducted under standardized administration guidelines, such as the \textit{Structured Interview Guide for the MADRS} (\textbf{SIGMA})~\cite{williams2008development}. Nevertheless, the rating remains dependent on clinician expertise and judgment, often introducing variability that can affect trial outcomes~\cite{lipsitz2004rater}. This challenge makes clinical trial interviews a particularly attractive setting for automated depression analysis.

Recent years have seen growing interest in automated methods for depression detection, with the aim of providing confidential and rapid triage for patients while supporting clinicians in their assessments~\cite{fisher2026language,zhang2026text,milintsevich2023towards}. Central to these efforts is the analysis of language~\cite{viduani2026from}, as language is a well‑established indicator of depression~\cite{trifu2024linguistic,segrin1990meta} and clinical interviews naturally unfold through speech. Accordingly, existing research has focused on increasingly sophisticated models that infer depression from text~\cite{liu2025enhanced}. However, much of this research has been conducted on online or non-clinical text, with comparatively little attention devoted to standardized clinical interviews used in depression trials~\cite{hengle2024still,squires2023deep,ji2022mentalbert,frohlich2018from}. Furthermore, researchers have primarily concentrated on binary classification model~\cite{zhang2025optimizing,bader2024detecting}, offering limited interpretability into individual symptoms that characterize depressive disorders.

\paragraph{Our original contributions.} In this work, we address these gaps by developing a large language model (LLM) pipeline for supporting the assessment of depression in patients enrolled in \textit{clinical trials}. Inspired by the structure of MADRS, we introduce the \madrsnospace, which orchestrates different LLMs to evaluates depressive symptoms directly from \textit{clinician}–\textit{patient} interview recordings. The pipeline first transcribes interview audio into text and segments the transcript according to the symptom items of MADRS. Then, for each item, it estimates symptom severity to support clinical rating. Finally, it compares these estimations against ratings provided by human clinicians and identifies instances where the human ratings may not fully comply with MADRS. %Finally, given ratings from human clinicians, it flags ratings that do not fully comply with MADRS.

We conduct extensive experiments across different LLMs to evaluate the performance of the complete pipeline. We validate it on  real-world clinical data, %trials,
comprising approximately 16,000 expert-rated MADRS instances. The results demonstrate strong agreement with expert assessments, achieving Spearman correlation of \textbf{0.867} on total MADRS scores.

\section{Background and Related Work}
Given the interdisciplinary nature of this work, we first summarize the clinical foundations underlying depression assessment and then review computational approaches to depression detection in NLP.

\paragraph{Clinical foundations} MADRS is one of the most widely used clinician-administered instruments for assessing depression severity in both clinical practice and drug‑development trials~\cite{montgomery1979new}. The scale consists of ten symptom items,\footnote{Apparent Sadness, Reported Sadness, Inner Tension, Reduced Sleep, Reduced Appetite, Concentration Difficulties, Lassitude, Inability to Feel, Pessimistic Thoughts, and Suicidal Thoughts} each rated on a severity scale from 0 to 6. 

Although well‑validated and sensitive to symptomatic change over time, the original MADRS offers no guidance on how clinicians should elicit information, leaving interview quality dependent on individual interviewing style and expertise. To reduce assessment variability across raters and study sites, structured administration protocols such as SIGMA were developed~\cite{williams2008development}. SIGMA standardizes interview administration by providing scripted questions and recommended follow-up probes for each symptom item. These guidelines ensure that clinicians systematically collect the information required for reliable symptom assessment and consistent scoring.

\begin{figure*}[!ht]
    \centering
    \includegraphics[width=0.87\linewidth]{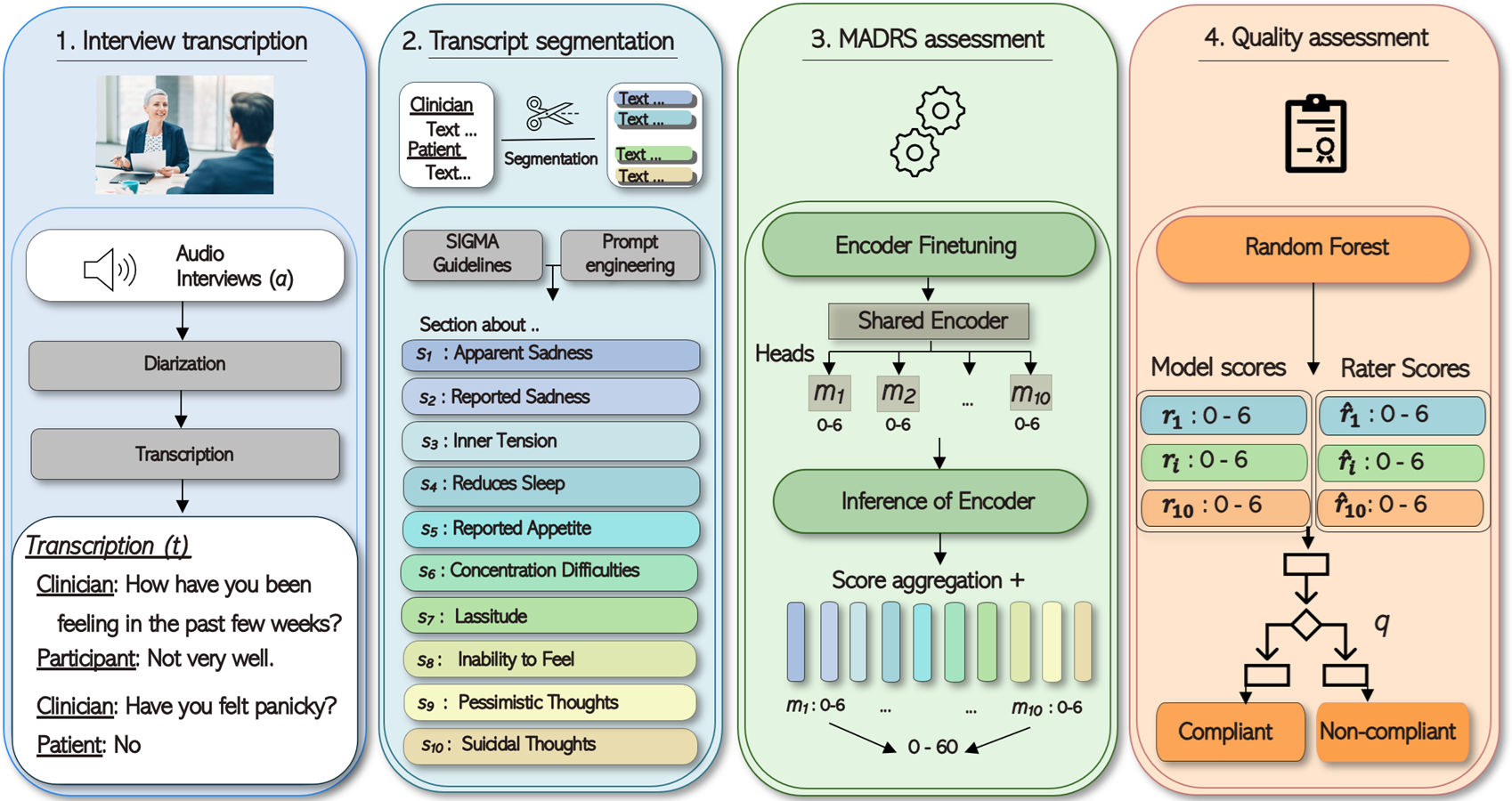}
    \caption{Overview of the \madrsnospace. The \madrs orchestrates four components (1) \textit{interview transcription}, (2) \textit{transcript segmentation}, (3) \textit{MADRS assessment}, and (4) \textit{quality assessment}.}
    \label{fig:agent}\vspace{-0.5cm}
\end{figure*}

\paragraph{Computational approaches} 
%FRANCESCO: this defends the choice of working with text
Research on depression detection spans multiple modalities, including audio‑based analyses~\cite{sardari2022audio}, video‑derived behavioral markers~\cite{mahayossanunt2023explainable}, %namunu2009video
and multimodal systems~\cite{shangguan2022dual}. %pampouchidou2016depression,makiuchi2019multimodal
Despite this advancements, text remains the most widely used data source due to data privacy concerns (\S~\ref{sec:ethical}), and the central role of language in clinical assessment. %FRANCESCO: Prior work has examined a variety of textual inputs, ranging from social‑media content such as Twitter~\cite{resnik2015beyond} and Facebook posts~\cite{schwartz2014towards} to student essays~\cite{resnik2013using}, screening and clinical interviews~\cite{rinaldi2020predicting,liu2025enhanced}.

%FRANCESCO: this to show that we are aware of other approaches
State‑of‑the‑art approaches rely on deep learning~\cite{squires2023deep}. Earlier work primarily uses CNNs and LSTMs to capture sentiment and semantic signals of depression~\cite{amna2022electronics}. More recent studies leverage LLMs for their unprecedented, natural-language capabilities~\cite{rinaldi2020predicting,liu2025enhanced,viduani2026from}. 

Prompt engineering is sometimes used to process interview transcripts as monolithic units~\cite{raganato2024madrs}; however, this approach requires long prompts and suffers from context overload, a problem referred to as ``\textit{lost-in-the-middle}''~\cite{liu2024lost}. Other approaches instead relies on multi-step prompting~\cite{kebe2026llamadrs,liu2025enhanced} to decompose interviews into fine-grained units---e.g., symptom-specific text segments~\cite{rinaldi2020predicting}, thematic blocks~\cite{zhao2025predicting}, and clinician–patient dialogue turns~\cite{lee2026plos}--- and then analyze each unit independently.\\

\textbf{Our work} builds upon these recent efforts. %KATA: recognizing the benefits of agentic approaches such as decomposing the task into smaller components, reducing context overload effects and its better suitability for multi-step, high-stakes tasks. 
Similar to~\citet{vail2026adapts}, we employ a LLM orchestration that leverages a mixture-of-models to solve an \textit{end-to-end} clinical assessment: unlike preliminary experiments relying on manually prepared transcripts~\cite{raganato2024madrs}, \madrs directly processes audio recordings, performs speaker-aware transcription, and identifies symptom-specific evidence. In contrast to~\citet{vail2026adapts}, \madrs is specifically designed to support standard MADRS scoring. 

Similar to~\citet{weber2025using}, \madrs employs a shared encoder with 9 item-specific heads for MADRS assessment. However, they focus on a subset of MADRS items, using real and synthetic German data. In contrast, \madrs is trained only on real English clinical interviews and predicts all 10 MADRS items. 

\citet{kebe2026llamadrs} represents another relevant comparison. However, their end-to-end reliance on larger-scale models (e.g., 70B parameters) introduces substantial computational costs in terms of both time and resources. In contrast, by leveraging different models, \madrs provides a practical trade-off between clinical utility, computational efficiency, and scalability. 

Finally, to the best of our knowledge, \madrs is the first computational solution specifically designed for depression clinical trials following SIGMA administration guidelines. Beyond supporting symptom severity estimation, \madrs enables quality monitoring by assessing the reliability of clinical ratings and identifying potentially non-compliant assessments.

\begin{table*}[t]
\centering
\small
\begin{tabular}{llcccc}
\toprule
\textbf{Component} & \textbf{Data split} & \textbf{Interviews} & \textbf{Utterances} & \textbf{Annotations} & \textbf{\# Annotations} \\
\midrule
Interview transcription &
Test &
8 &
2,642 &
Transcripts &
2,642 \\

\midrule

Transcript segmentation &
Test &
14 &
3,970 &
10 SIGMA sections &
$3,970\times10$ \\

\midrule

MADRS assessment &
Train &
1100 &
371,715 &
\multirow{4}{*}{10 MADRS scores (0--6)} &
$1100\times10$ \\

&
Dev &
251 &
87,364 &
&
$251\times10$ \\

&
Test &
251 &
86,789 &
&
$251\times10$ \\

%&
%\textit{Out}-Test &
%253 &
%96,635 &
%&
%$253\times10$ \\

\midrule

Quality assessment &
Train &
305 &
119,017 &
\multirow{2}{*}{$\{$Compliant, Non-compliant$\}$} &
$305\times2$ \\

&
Test &
305 &
117,324 &
&
$305\times2$ \\\bottomrule
\end{tabular}
\caption{\textbf{Datasets}: for each component and data split, we report the number of interviews (duration: \textasciitilde40 min), the total number of utterances across interviews, the annotation format, and the total number of annotations.}
\label{tab:datasets}
\end{table*}

\section{The \madrs}%What is the proposed method?
Let $a$ denote the audio recording of a clinician--patient interview conducted within a depression clinical trial according to the SIGMA guidelines. The problem addressed by the \madrs is to support the depression assessment of the patient across the ten symptom items of the MADRS scale, denoted by $\mathcal{M}={m_1,\ldots,m_{10}}$.

The \madrs is designed as a \textit{computational clinician} grounded in standardized clinical interviewing procedures. It is implemented as a directed acyclic graph comprising an \textit{orchestration} of four sequential components, where each stage operates on the outputs of preceding stages. The orchestration consists of: (1) \textit{interview transcription}, (2) \textit{transcript segmentation}, (3) \textit{MADRS assessment}, and (4) \textit{quality assessment}. An overview of the proposed framework is shown in Figure~\ref{fig:agent}.

\paragraph{Interview transcription} This component coverts the audio recording $a$ into a textual transcript $t$. The resulting transcript contains a chronological sequence of clinician and patient utterances and serves as the input to subsequent processing stages.

To generate transcripts, we follow an established pipeline used in related research~\cite{costin2026speaker,desai2024advancing,weber2025using}. Specifically, we employ an open-source deep learning toolkit (i.e., \texttt{pyannote.audio 4.0};~\citealp{bredin2019audio}) for voice activity detection and speaker diarization, combined with an automatic speech recognition (ASR) system (i.e., \texttt{Whisper Medium};~\citealp{radford2023robust,desai2024advancing}) to transcribe the audio recordings.

\paragraph{Transcript segmentation} Clinical interviews are organized around the MADRS items $\mathcal{M}$. This component decompose a transcript $t$ into 10 symptom-specific segments $S={s_1,\ldots,s_{10}}$, where each segment $s_i$ contains the interview utterances relevant to the assessment of the corresponding MADRS item $m_i$. The resulting representation aligns the interview with the structure of the clinical instrument, enabling subsequent components to operate on localized symptom-specific evidence rather than the transcript as a whole.

To segment transcripts, we employ a LLM (i.e., \texttt{GPT-4.1}) guided by the corresponding SIGMA administration guidelines. For each MADRS item, the model receives the complete transcript together with the item-specific SIGMA description and it is instructed to identify the utterances relevant to the target symptom using a zero-shot prompting strategy. This procedure is repeated independently for all ten MADRS items (Appendix~\ref{app:transcript-segmentation}). %CAMERA READY: Additional details on the used prompt are provided in .

\paragraph{MADRS assessment} This is the core component of \madrsnospace, responsible for estimating the severity of each MADRS symptom. For each symptom item $m_i$, the component analyzes the corresponding segment $s_i$ and predicts a severity rating $r_i \in \{0,\ldots,6\}$, where 0 indicates the absence of the symptom and 6 corresponds to the highest level of severity defined by the MADRS guidelines.

To estimate symptom severity, we employ a Transformer encoder model fine-tuned for multi-task ordinal regression~\cite{baly2019multitask,li2006ordinal}. The model processes each segment $s_i$ independently and predicts the corresponding MADRS ordinal rating. Specifically, a shared encoder first maps the input text into a latent representation, which is subsequently processed by item-specific prediction heads corresponding to the ten MADRS symptoms. Each head models the ordinal structure of MADRS ratings through a sequence of threshold decisions, enabling the prediction of scores on the discrete 0--6 scale while explicitly accounting for their ordered nature.

\paragraph{Quality assessment} In clinical trials, \textit{reviewer} teams routinely monitor clinical \textit{raters} to ensure their adherence to MADRS scoring guidelines. This component is designed to support this process by detecting interviews where the \textit{rater} assessment is not consistent with expected clinical practice. 

Given the MADRS scores ${\hat{r}_1,\ldots,\hat{r}_{10}}$ assigned by the clinical \textit{rater} as additional input, this component compares them with the \textit{reference} scores ${r_1,\ldots,r_{10}}$. The reference scores can either be generated directly by the \madrs\ (when a review by the \textit{reviewer} team is not available), or provided/refined by the \textit{reviewer} team following their assessment. The component then predicts a binary quality label, $q \in \{$Compliant, Non-compliant$\}$, indicating whether the \textit{rater} scores are considered reliable or potentially problematic.

To estimate interview compliance, we train a Random Forest binary classifier using the absolute differences, $|r_i - \hat{r}_i|$ for $i \in \{1,\ldots,10\}$, as input features. As additional features, we include the \textit{total} and the \textit{mean} of the absolute differences.

\section{Experimental setup}
We evaluate \madrs on a collection of interviews annotated with MADRS scores from both clinical \textit{raters} and \textit{reviewer} team at the participating clinical sites. Since \textit{interview transcription} and \textit{transcript segmentation} constitute essential pre-processing steps for the subsequent clinical assessments, we conduct targeted validation analyses to verify the expected quality-level of the speech recognition pipeline.\footnote{Larger-scale evaluation of these components would require extensive review of interview audio recordings and the generation of additional ground-truth labels, an effort which is outside the scope of this work.}

\subsection{Dataset}
Our dataset consists of 1,602 %clinician–patient 
interviews administered according to SIGMA. Each interview contains approximately 40 minutes of audio. Trained \textit{raters} in private clinics conducted real-time interviews and assigned 10 MADRS scores during each assessment. A \textit{reviewer} team evaluated the audio recordings of these interviews and assigned an additional set of MADRS scores to assess scoring consistency. They then labeled a subset of the interviews as \textit{compliant} or \textit{non-compliant} based on discrepancies between the two sets of MADRS scores. Ground-truth collection and clinical training procedures were defined by the study protocol and are thus outside the scope of this work (Appendix~\ref{app:ira}). 

We randomly selected two subsets of interviews to assess \textit{interview transcription} and \textit{transcript segmentation}. For \textit{interview transcription}, we manually transcribed 8 interviews (2,642 utterances) to serve as the reference standard. For \textit{transcript segmentation}, we annotated 14 interviews (3,970 utterances) by assigning each utterance to one of the 10 SIGMA interview sections, each corresponding to a MADRS item. 

For \textit{madrs assessment}, we randomly partitioned interviews into Train, Dev, and Test sets using an 80/10/10 split\footnote{No patient appears in more than one set.} and considered ground truth provided by the \textit{reviewer} team. For \textit{quality assessment}, the subset annotated with quality labels was smaller and exhibited class imbalance. We therefore conducted 10-fold cross-validation using random 50/50 Train/Test splits while preserving the class distribution. A summary of our dataset is available in Table~\ref{tab:datasets}.

\subsection{Task setting}
\paragraph{Interview transcription} 
We conduct a quality assessment by computing the similarity between the automatically generated transcripts and the manually curated references. First, we use the temporal Dice metric to align utterances across the two transcript versions. Then, we quantify lexical similarity using the average Jaccard index over the aligned utterances and semantic similarity using the average cosine similarity between sentence embeddings\footnote{\small \texttt{sentence-transformers/all-MiniLM-L6-v2}} of the aligned utterances (Appendix~\ref{app:transcription}). %CAMERA READY: Further details are available in Appendix~\ref{app:transcription}.

\paragraph{Transcript segmentation} We evaluate this component as a multi-class classification task, where the objective is to assign each utterance to one of the 10 SIGMA sections. Performance is measured using one-vs-all F1-score for each SIGMA section and weighted F1-score as an overall metric. 

We compare one open-weight LLM, \texttt{Llama 3.3 70B}, with two closed-source LLMs, \texttt{GPT-4.1} and \texttt{Claude Sonnet 4.5}. These models were selected because their large context windows (128K, 1M, and 200K tokens, respectively) enable processing of long clinical transcripts. To minimize variability due to the non-deterministic nature of LLMs, decoding is performed with temperature set to 0. Each experiment is repeated three times, and the average result across runs is considered.

\begin{figure}[t]
    \centering
    \includegraphics[width=0.95\columnwidth]{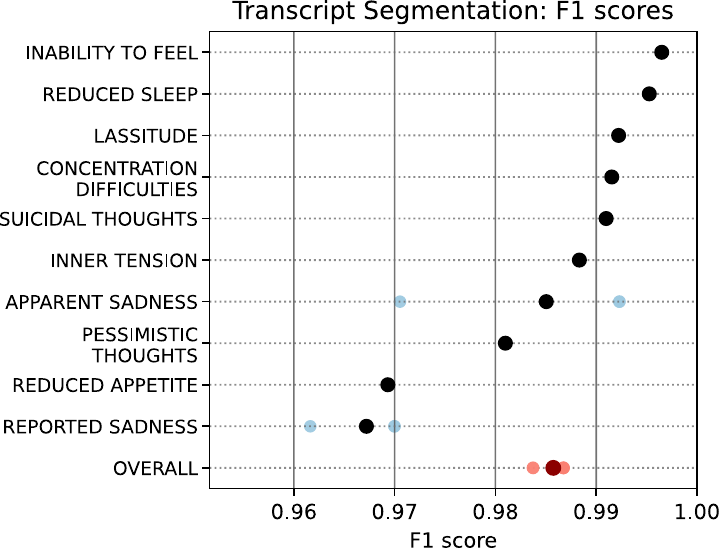}
    \caption{One-vs-all F1 scores for \textit{transcript segmentation} across three runs. Light points: individual runs; dark points: mean. Red: overall performance.}
    \label{fig:segmentation}\vspace{-0.7cm}
\end{figure}

\paragraph{MADRS assessment} We evaluate this component as a ranking task by comparing our prediction against the scores of the \textit{reviewer} team at both the item and total score level. Performance is measured using complementary metrics: Spearman correlation, Mean Absolute Error (MAE), and Accuracy@1. Spearman correlation assesses ranking quality. MAE measures the absolute error in predicted overall MADRS scores. Accuracy\@1 reports the proportion of predictions within a $\pm 1$ point tolerance of the expert ratings. 

We compare fine-tuned encoders across three settings: \textbf{(i)} pretraining, \textbf{(ii)} architecture, and \textbf{(iii)} training loss. For\textbf{ (i)}, we compare general-purpose and clinically adapted encoders jointly fine-tuned across all items using an ordinal regression loss. For \textbf{(ii)}, we compare the best-performing model from \textbf{(i)} with independent models fine-tuned separately for each MADRS item. For \textbf{(iii)}, we compare the same best-performing model under ordinal regression and categorical classification loss (Appendix~\ref{app:encoder-fine-tuning}). All models are trained for up to 10 epochs with early stopping based on validation item-level MAE. Specifically, we evaluate \texttt{RoBERTa} ~\cite{zhuang2021robustly}, \texttt{ModernBERT}~\cite{warner2025smarter}, \texttt{MentalBERT}~\cite{ji2022mentalbert}, \texttt{ClinicalBERT} ~\cite{huang2020clinicalbert}, and \texttt{PubMedBERT} ~\cite{gu2022domain}.

Depression datasets are typically confidential and therefore not publicly available (\S~\ref{sec:ethical}). As a result, direct comparison with state-of-the-art results is not feasible. To provide a relevant baseline with state-of-the-art, we additionally evaluate decoder LLMs used for \textit{transcript segmentation}, following the zero-shot LLAMADRS prompting strategy proposed by~\citet{kebe2026llamadrs}.

\paragraph{Quality assessment} We evaluate this component as a binary classification task by comparing our predictions against the quality labels of the \textit{reviewer} team. Given the imbalanced nature of the data and the importance of negative examples in this screening setting (\S~\ref{sec:limitation}), we use the Matthews Correlation Coefficient (MCC) as %evaluation 
metric and report True Positives (TP), True Negatives (TN), False Positives (FP), and False Negatives (FN). 

We evaluate \madrs both when the reference scores ${r_1,\ldots,r_{10}}$ are provided by the \textit{reviewer} team, and when they are directly generated by the top-performing \textit{encoder} and \textit{decoder} from \textit{MADRS assessment}. 

When reference scores are directly generated\footnote{To prevent information leakage, MADRS reference scores were generated using split-specific fine-tuned models that excluded the corresponding quality assessment interviews (see Appendix~\ref{app:quality-assessment}).}, we account for the fact that MADRS predictions are not error-free. For each MADRS item, we thus investigate the use of a \textit{noise quantile threshold} estimated from the differences between the generated scores and the scores provided by the \textit{reviewer} team. We then use this threshold when comparing the \textit{rater} scores with the generated scores, so that only disagreements larger than the expected model error are counted. This reduces the risk of attributing model error to the clinical \textit{rater} (Appendix~\ref{app:quality-assessment}). %CAMERA-READY: Detailed formulation is provided in the Appendix~\ref{app:quality-assessment}.

To better contextualize task difficulty, we include simple baselines, namely Random, All-1s, and All-0s. Additionally, we report performance using a simple threshold classifier tuned on the total gap (sum of the item differences) over the Train set.

\subsection{Evaluation results}%How well does it work?
\paragraph{Interview Transcription} Our validation suggests strong alignment at the utterance level, reaching an average temporal Dice score of 0.852 between generated and manually curated utterances. On the aligned utterances, we observe a high similarity, with average Jaccard index of 0.749 and a average embedding similarity of 0.814. This analysis suggests substantial overlap in both wording and meaning, supporting the use of generated transcripts for the subsequent components.

%\paragraph{Interview Transcription} Our evaluation suggests strong alignment at the utterance level, reaching an average temporal Dice score of 0.85 between generated and manually curated utterances. On the aligned utterances, we observe high lexical agreement, with an average Jaccard index of 0.75, indicating substantial overlap in word usage between generated and reference transcripts. We also observe high semantic similarity, with an average score of 0.81, which further confirms that the generated transcripts preserve the overall meaning of the reference content, even in cases where exact lexical matches are limited.

\begin{figure}[!ht]
    \centering
    \includegraphics[width=\columnwidth]{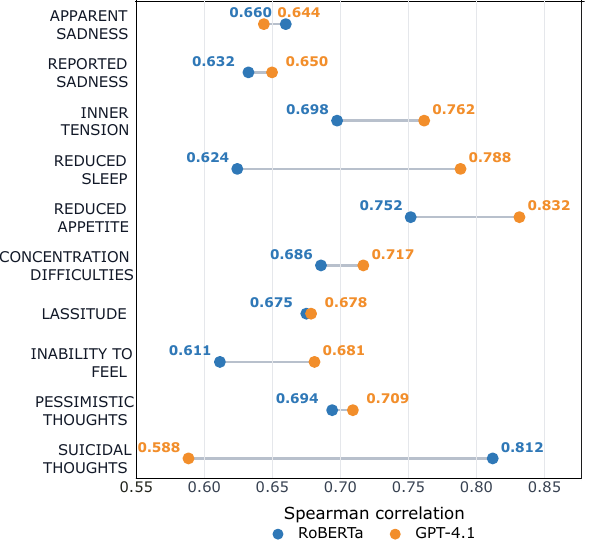}
    \caption{Spearman correlation of the best-performing encoder (\texttt{RoBERTa}) and decoder (\texttt{GPT-4.1}) models for \textit{MADRS assessment} on individual MADRS items.}
    \label{fig:item-spearman}\vspace{-0.3cm}
\end{figure}

\paragraph{Transcript segmentation} The \textit{overall} results across three independent evaluation runs are reported in Table~\ref{tab:segmentation}. Our results show that closed-source models outperform the open-weight model, with \texttt{GPT-4.1} achieving slightly higher performance and therefore being selected as the reference model. Figure~\ref{fig:segmentation} reports the performance of \texttt{GPT-4.1} across the individual MADRS items. Overall, the segmentation module achieved consistently strong performance, with one-vs-all F1 scores exceeding 0.94 for all MADRS items and an overall weighted F1 above 0.98. Performance was highly stable across runs (\textit{std} from 0.005 to 0.01), as evidenced by the close clustering of individual run-level scores around their corresponding means, indicating that the segmentation procedure is robust to residual variability in LLM inference. The high performance suggests that symptom-specific segmentation is a well-structured task in standardized SIGMA interviews. The predefined interview structure creates substantial lexical and semantic overlap between the symptom descriptions supplied in the prompt and the corresponding interview content (Appendix~\ref{app:transcript-segmentation}). This redundancy may simplify symptom identification, aligning with recent evidence that repeated contextual information can improve the performance of non-reasoning LLMs~\cite{leviathan2025prompt}.

\begin{table}[!ht]
\resizebox{\columnwidth}{!}{%
\begin{tabular}{lllll}
\textbf{weigh. F1 score} & Run1 & Run2 & Run3 & \textbf{Avg.} \\ \hline
\texttt{GPT-4.1}   &  \textbf{0.984} & \textbf{0.981} & \textbf{0.984} & \textbf{0.983}  \\
\texttt{Llama 3.3 70B} &  0.654   & 0.609 & 0.582 & 0.615\\
\texttt{Claude Sonnet 4.5}  & 0.978 & 0.977 & 0.978 & 0.978 \\ \hline
\end{tabular}}
\caption{Weighted F1 for \textit{transcript segmentation}.}
\label{tab:segmentation}
\end{table}

\begin{table*}[t]
\centering
\small
\setlength{\tabcolsep}{6pt}
\begin{tabular}{lcccc}
\toprule
Model & Setting & MAE & Accuracy@1 & Spearman \\%FRANCESCO: Adding total is confusing here. Let's clarify in text and caption.
\midrule
RoBERTa & (i-iii) \ - \hfill \textit{ general purpose} & \textbf{2.956} & \textbf{0.895} & \textbf{0.867} \\
RoBERTa - {\small Independent fine-tuning} & (ii) \hspace{0.18cm}  \ - \hfill \textit{general purpose} &  3.493 & 0.860 & 0.800 \\
RoBERTa - {\small Categorical multiclass loss} & (iii) \hspace{0.1cm}  \ - \hfill \textit{general purpose} & 3.039 & 0.900 & 0.850 \\
ModernBERT & \ (i) \hspace{0.2cm}  \ - \hfill \textit{general purpose} & 4.200 & 0.842 & 0.716 \\
MentalBERT & \ (i) \hspace{0.2cm} \ - \ \hfill \textit{clinically adapted} & 3.288 & 0.879 & 0.842 \\
PubMedBERT & \ (i) \hspace{0.2cm} \ - \ \hfill\textit{clinically adapted} & 4.161 & 0.836 & 0.756 \\
ClinicalBERT & \ (i) \hspace{0.2cm} \ - \ \hfill\textit{clinically adapted} & 4.322 & 0.832 & 0.754 \\
\midrule
\textbf{Baselines:}~\citet{kebe2026llamadrs} & Prompt-engineering & MAE & Accuracy@1 & Spearman \\
\midrule
\texttt{GPT-4.1} & LLAMADRS & 4.044 & 0.813 & 0.849 \\
\texttt{Claude Sonnet 4.5} & LLAMADRS & 4.080 & 0.814 & 0.848 \\
\texttt{Llama 3.3 70B} & LLAMADRS & 4.133 & 0.811 & 0.851 \\
\bottomrule
\end{tabular}
\caption{Summary performance for \textit{MADRS assessment}: MAE and Spearman are computed over total MADRS scores. For Accuracy@1, we report the average result per item, as the metric would be too aggressive at the total level. For \textbf{(i)}, encoders are all jointly fine-tuned across all items using an ordinal regression loss.}
\label{tab:madrs-results}\vspace{-0.5cm}
\end{table*}

\vspace{-0.3cm}

\paragraph{MADRS Assessment}
Table~\ref{tab:madrs-results} summarizes the results of our evaluation considering the total MADRS scores. Overall, \texttt{RoBERTa} achieves the best performance, with an MAE of 2.956, an avg. Accuracy@1 of 0.895, and a Spearman correlation of 0.867. Baseline LLMADRS also achieve strong performance, although it consistently underperform the fine-tuned encoder models. %In particular, we observe that closed-source models outperform their open-weight counterpart.

Figure~\ref{fig:item-spearman} reports the %Spearman 
correlation for individual %MADRS 
items for %achieved by 
fine-tuned \texttt{RoBERTa} and %one of the strongest 
a decoder baselines, \texttt{GPT-4.1}. Specifically, we observe that performance varies across items, and that \texttt{GPT-4.1} outperform \texttt{RoBERTa} on most items. \texttt{RoBERTa} achieves higher correlations than \texttt{GPT-4.1} only for \textit{Apparent Sadness} and \textit{Suicidal Thoughts}, with the largest gap observed for the latter. We hypothesize that the sensitive nature of the item may interact with safety guardrails~\cite{dong2025safeguarding}, potentially leading to more conservative predictions and reduced performance. Despite weaker per-item correlations, we shown in Table~\ref{tab:app_acc} and~\ref{tab:app_mae}, that \texttt{RoBERTa} outperforms all baselines in terms of MAE and Accuracy@1. This translates into higher performance on the total MADRS score, suggesting the aggregation of item predictions can mitigate the overall measurement noise.

\begin{figure}[t]
    \centering
    \includegraphics[width=\linewidth]{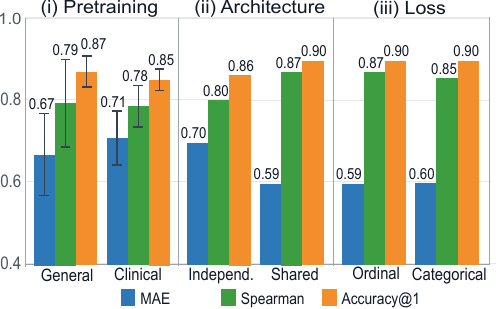}
    \caption{Performance for \textit{MADRS assessment} over total MADRS scores in setting \textbf{(i-iii)}.}
    \label{fig:madrs-experiment-categories}\vspace{-0.5cm}
\end{figure}

\begin{figure}[t]
    \centering
    \includegraphics[width=0.95\linewidth]{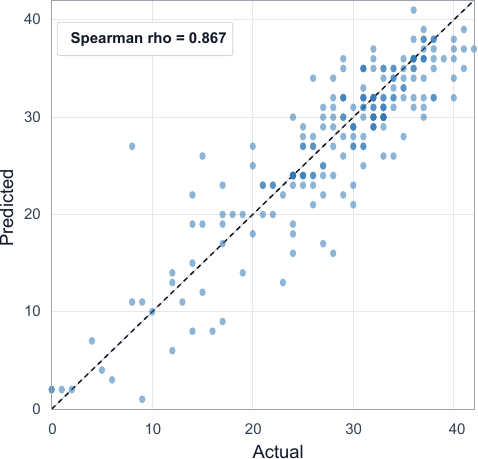}
    \caption{Spearman for \textit{MADRS assessment} over total MADRS scores using fine-tuned \texttt{RoBERTa}.}
    \label{fig:madrs-spearman}\vspace{-0.4cm}
\end{figure}

Figure~\ref{fig:madrs-experiment-categories} and Table~\ref{tab:madrs-results} summarize our evaluation over pretraining \textbf{(i)}, architecture \textbf{(ii)}, and loss \textbf{(iii)}. For pretraining \textbf{(i)}, the %observed 
performance differences are relatively small; However, in line with recent observations \cite{vishwanath2026general}, general-purpose \texttt{RoBERTa} achieves better point estimates than its clinical counterparts. Among the evaluated models, \texttt{RoBERTa} achieves the strongest correlation (Figure~\ref{fig:madrs-spearman}) and performance (Appendix~\ref{app:additional-results}) and is therefore selected for the experiments \textbf{(ii)} and \textbf{(iii)}.

Considering \textbf{(ii)}, shared fine-tuning consistently outperforms independent fine-tuning. This suggests that a shared model benefits from common patterns across MADRS items. For example, negative responses to key symptom questions (e.g., ``No'' for symptom presence) are informative across multiple items, as they are consistently associated with lower scores. Similarly, recurring severity anchors such as \textit{normal} versus \textit{reduced} provide consistent signals that generalize across different symptoms.

Considering \textbf{(iii)}, we observe that the categorical loss yields nearly identical performance, while the ordinal loss achieves slightly higher Spearman correlation on the total score.

\begin{figure}[t]
    \centering
    \includegraphics[width=\linewidth]{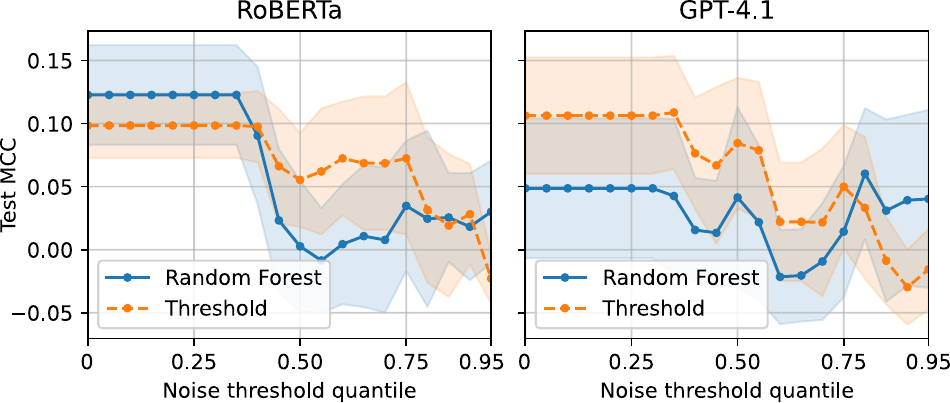}
    \caption{Spearman for \textit{MADRS assessment} over total MADRS scores using fine-tuned \texttt{RoBERTa}.}
    \label{fig:quality-assessment}\vspace{-0.3cm}
\end{figure}

\paragraph{Quality assessment}
We report in Figure~\ref{fig:quality-assessment} the results of our evaluation while varying the quantile threshold used to discount expected prediction error when generated scores are used as \textit{reference} scores. Overall, lower levels of discounting lead to higher MCC, with the best performance obtained when no discount is applied. Performance progressively decreases as the model becomes more tolerant to prediction errors.  

Table~\ref{tab:quality-assessment} reports the result of our evaluation when no discount is applied. The highest performance is achieved when using MADRS scores provided by the \textit{reviewer} team as \textit{reference} scores, reaching an MCC of 0.704. When using MADRS scores automatically generated by \madrs\ as reference scores, performance remains above the baselines but decreases due to error propagation from the automatic predictions. In particular, we obtain a MCC of 0.123 when using predictions generated by fine-tuned \texttt{RoBERTa}, and a MCC of 0.061 when using predictions generated by \texttt{GPT-4.1}. This suggests that, although reviewer scores remain the preferred option, \texttt{RoBERTa} scores can still provide a useful signal for identifying \textit{non-compliant} ratings. The superior performance of \texttt{RoBERTa} is consistent with its higher Accuracy@1 on \textit{MADRS assessment}.

Compared with the thresholding baseline approach, the Random Forest classifier achieves comparable performance when reviewer scores are used as references (MCC 0.704 vs.\ 0.713).  However, the comparison differs for RoBERTa and GPT-4.1. With the more accurate \texttt{RoBERTa} predictions, the Random Forest consistently outperforms the baseline, suggesting that the quality label $q$ depends on structured patterns of item-level discrepancies rather than only on their overall magnitude. In contrast, when \texttt{GPT-4.1} predictions are used, the threshold baseline performs better than the Random Forest. For \texttt{GPT-4.1}, the confidence intervals are substantially wider for both methods, reflecting the higher uncertainty introduced by the lower accuracy of the underlying MADRS predictions. % Specifically, when reviewer scores are used, the MCC decreases from 0.806 to 0.766; similarly, when scores from \texttt{RoBERTa} and \texttt{GPT-4.1} are used, the MCC decreases to 0.061 and 0.022, respectively. 
%Figure~\ref{fig:quality-assessment} shows that this trend is generally consistent across different quantile settings.

\begin{table}[t]
\centering
\small
\resizebox{\columnwidth}{!}{%
\begin{tabular}{lccccc}
\toprule
\textbf{Ran. For.} & MCC & TN & TP & FP & FN\\
\midrule
\textit{reviewer} &
\textbf{0.704 $\pm$ 0.1} &
24.1 $\pm$ 1.0 &
263.0 $\pm$ 3.9 &
5.9 $\pm$ 1.0 &
12.0 $\pm$ 3.9 \\
\texttt{RoBERTa} &
0.123 $\pm$ 0.0 &
8.2 $\pm$ 2.8 &
239.4 $\pm$ 12.0 &
21.8 $\pm$ 2.8 &
35.6 $\pm$ 12.0 \\
\texttt{GPT-4.1} &
0.049 $\pm$ 0.1 &
7.6 $\pm$ 3.1 &
223.5 $\pm$ 12.7 &
22.4 $\pm$ 3.1 &
51.5 $\pm$ 12.7 \\
\midrule
\textbf{Thresh.} & MCC & TN & TP & FP & FN\\
\midrule
\textit{reviewer} &
\textbf{0.713 $\pm$ 0.1} &
24.4 $\pm$ 1.3 &
263.5 $\pm$ 3.3 &
5.6 $\pm$ 1.3 &
11.7 $\pm$ 3.3 \\
\texttt{RoBERTa} &
0.098 $\pm$ 0.0 &
8.6 $\pm$ 6.0 &
120.2 $\pm$ 49.3 &
21.4 $\pm$ 6.0 &
154.8 $\pm$ 49.3 \\
\texttt{GPT-4.1} &
0.106 $\pm$ 0.0 &
9.2 $\pm$ 3.1 &
133.1 $\pm$ 14.9 &
20.8 $\pm$ 3.1 &
141.9 $\pm$ 14.9 \\
\midrule
\textbf{Baselines} & MCC & TN & TP & FP & FN\\
\midrule
Random & 0.03 & -- & -- & -- & -- \\
All-1s & 0 & 30 & 0 & 0 & 275 \\
All-0s & 0 & 0 & 275 & 30 & 0 \\
\bottomrule
\end{tabular}}
\caption{MCC for \textit{quality assessment}. Results are reported as mean $\pm$ standard dev. across 10 folds.}
\label{tab:quality-assessment}
\vspace{-0.6cm}
\end{table}

\section{Conclusion}
In this work, we introduced \madrsnospace, an end-to-end LLM pipeline that supports the assessment of depression in clinical trials. \madrs\ is specifically designed for clinical interviews in which the interviewing clinician assesses depression using the MADRS scale and follows the SIGMA guidelines.

\madrs\ orchestrates four components: \textit{interview transcription}, which converts clinical interview audio into structured speaker-aware transcripts; \textit{transcript segmentation}, which aligns utterances with the ten MADRS symptom items using structured SIGMA guidelines; \textit{MADRS assessment}, which estimates depression severity for each MADRS item; and \textit{quality assessment}, which identifies problematic clinical ratings.

We evaluated each component separately using real-world clinical trial data. First, we performed sanity checks to ensure the reliability of the generated structured representations of clinical interviews. We then evaluated the MADRS assessment and quality assessment components. Our results show strong agreement with expert reviewer ratings, reaching a Spearman correlation of 0.867 on total MADRS scores. Furthermore, we show that \madrs\ can identify potentially \textit{non-compliant} ratings using either generated MADRS predictions or reviewer-provided scores, although reviewer scores remain the preferred option. 

This is consistent with the intended role of \madrs\ as a supportive tool for clinicians rather than a replacement for clinical judgment or decision-making. Overall, \madrs\ represents a step toward more consistent, scalable, and interpretable psychiatric assessment workflows by supporting clinicians in the evaluation of depression severity and rating quality.

\section*{Limitations}\label{sec:limitation}
While the proposed \madrs demonstrates promising results for automated MADRS assessment, the following limitations should be considered when interpreting the findings.

\paragraph{Modality mismatch}
Human clinicians typically assess depression severity using in-person interviews conducted in real time, or audio and video recordings. In contrast, the \madrs operates exclusively on textual transcripts for \textit{MADRS} and \textit{quality assessment}. 

On the one hand, this design choice is motivated by practical considerations, including privacy, confidentiality, and data-sharing constraints that frequently arise in clinical research settings. On the other hand, restricting the analysis to transcripts removes access to paralinguistic and behavioral signals, such as prosody, intonation, pauses, facial expressions, and body language, which may contribute to clinical assessments. 

Consequently, there is an inherent modality mismatch between human and automated evaluations. Although the proposed framework achieves substantial agreement with expert \textit{reviewers}, its performance should be interpreted in light of this discrepancy. Future work could investigate multimodal approaches that integrate acoustic and visual information alongside textual evidence.

\paragraph{Transcription quality}
The \madrs relies on automatically generated transcripts, making it inherently sensitive to transcription errors. Although modern ASR systems generally achieve high accuracy and our evaluation indicates strong transcription quality, errors may still arise due to poor audio conditions, overlapping speech, speaker diarization failures, or model hallucinations. In clinical interviews, even subtle transcription inaccuracies can alter symptom descriptions and affect downstream assessments.

While aggregating predictions across multiple MADRS items may reduce the impact of isolated errors, transcription quality remains a fundamental dependency of the overall pipeline. Improvements in downstream modeling cannot fully compensate for missing or incorrectly transcribed information. Consequently, the reported results should be interpreted relative to the quality of the initial recordings and subsequent transcripts.

\paragraph{Imbalanced data in \textit{quality assessment}} The dataset used for quality assessment is highly imbalanced. While this poses challenges for model training and evaluation, the observed class distribution is representative of real-world screening conditions. Clinical \textit{raters} are trained to administer assessments in a standardized and consistent manner, which contributes to the relatively low prevalence of negative cases.

Constructing a reliable ground truth requires the sampling and independent validation of a large number of interviews in order to identify a relatively small number of erroneous instances, making the annotation process both costly and time-consuming. In our dataset, only 60 out of 610 interviews (approximately 407 hours of audio) are labeled as problematic according to MADRS-based scoring.

Despite this imbalance, the setting is of practical importance, as automated \textit{quality assessment} has the potential to substantially support and scale human expertise in clinical trial environments. As a first work based on real-world data, our work contributes toward the development of supportive tools for clinical trials. As a future direction, we envision a \textit{human-in-the-loop} framework in which the dataset can be progressively expanded through additional expert annotations, and model performance iteratively improved through feedback-driven refinement.

%\paragraph{Inter-annotator agreement}
%MADRS assessment is an inherently subjective task that requires specialized training and adherence to detailed rating guidelines. Even among experienced raters, complete agreement is rarely achieved. The inter-rater variability observed in our data reflects the presence of genuine ambiguity in symptom interpretation and severity assessment.

%This variability imposes an upper bound on the agreement achievable by any automated system trained to reproduce human judgments. As a result, discrepancies between model predictions and reference ratings do not necessarily indicate model failure; they may instead reflect differences among plausible clinical interpretations. We therefore view the proposed framework as a decision-support tool intended to assist and standardize expert review rather than replace human judgment.

\section{Ethical Considerations}\label{sec:ethical}
This work involves the analysis of sensitive mental health data collected as part of clinical trials. All interview recordings and associated annotations were de-identified prior to analysis in accordance with the data governance procedures of the sponsoring studies. No personally identifiable information was used during model development or evaluation.

To further reduce privacy risks, the proposed framework operates primarily on textual transcripts rather than raw audio or video recordings. While this design choice introduces a modality mismatch with human assessments, it limits exposure to potentially identifiable acoustic and visual information, thereby facilitating privacy-preserving analysis of clinical interviews.

The proposed \madrs is intended as a decision-support tool and not as a replacement for trained clinical raters. Automated assessments of depression severity may be affected by transcription errors, model biases, and ambiguity inherent to psychiatric evaluation. Consequently, system outputs should be reviewed by qualified professionals and should not be used as the sole basis for clinical or treatment decisions.

Due to confidentiality agreements and regulatory restrictions governing clinical trial data, neither the underlying datasets nor the trained models can be publicly released. To support transparency and reproducibility, we release model configurations, prompting templates, and synthetic examples that replicate the structure of the evaluation pipeline without exposing confidential material. %TODO: ?? We will additionally provide mock interviews and corresponding annotations to facilitate future methodological comparisons and benchmarking.

\section*{Acknowledgments}
This work was supported by the NeuRev project, funded by Clario, part of Thermo Fisher Scientific. AI coding tools were used to support software development; all generated code suggestions were reviewed, adapted, validated, and integrated by the authors, who remain fully responsible for the final implementation. We thank Matthew Agard, and Rachel Alexander for their constructive discussions, brainstorming, feedback, and support throughout this project. We also thank Milos Ivankovic and Cuong Lai for developing the backend and user interface of the pipeline, which eventually led our scientific research toward a usable product; and Volodymyr Pozniak for facilitating access to the necessary infrastructure. Finally, we are grateful to Barbara Echevarria and Mark Opler for their leadership of the Clinical Science team and their efforts in curating and collecting the foundational data that supported this work.

\bibliography{custom}

\appendix

\section{Transcription}\label{app:transcription}
\subsection{Transcription quality}
%For all possible pairs of manually transcribed utterances ${\hat{u}_i}$, and automatically generated transcription utterances ${{u}_i}$, temporal alignment was determined using the Dice coefficient, defined as \begin{equation}
%\mathrm{Dice}(u_i, \hat{u}_i) =
%\frac{2 \lvert t(u_i) \cap t(\hat{u}_i) \rvert}
%{\lvert t(u_i) \rvert + \lvert t(\hat{u}_i) \rvert}
%\end{equation} where ${{t}({u}_i)}$ denotes the temporal span of an utterance ${{u}_i}$. This metric was used to identify the best-matching utterance pairs in time. Subsequently, lexical similarity was quantified using the Jaccard index, and semantic similarity was computed as the average utterance cosine similarity between sentence embeddings derived from Sentence Transformers.

We conduct a quality assessment by computing the similarity between the automatically generated transcripts and the manually curated references. 
For all possible pairs of manually transcribed utterances ${\hat{u}_i}$, and automatically generated transcription utterances ${{u}_i}$, temporal alignment was determined using the Dice coefficient, defined as 
\[
\mathrm{Dice}(u_i, \hat{u}_i) =
\frac{2 \lvert t(u_i) \cap t(\hat{u}_i) \rvert}
{\lvert t(u_i) \rvert + \lvert t(\hat{u}_i) \rvert}
\] where ${{t}({u}_i)}$ denotes the temporal span of an utterance ${{u}_i}$. The optimal temporal matching is defined as the pairing that maximizes the Dice coefficient.

Lexical similarity was quantified using the average Jaccard index between utterances $u_i$ and ${\hat{u}_i}$; where Jaccard is defined as:
\[
\mathrm{Jac.}(u_i, \hat{u}_i) =
\frac{ \lvert words(u_i) \cap words(\hat{u}_i) \rvert}
{ \lvert words(u_i) \cup words(\hat{u}_i) \rvert} \ .
\]

Similarly, semantic similarity was quantified as the average cosine similarity between sentence embeddings derived from Sentence Transformers; where Cosine is defined as:
\[
\cos(u_i, \hat{u}_i) = \frac{u_i \cdot \hat{u}_i}{\|u_i\| \, \hat{u}_i\|}  \ .
\]

\subsection{Transcript segmentation}\label{app:transcript-segmentation}
We adopt a zero-shot prompt engineering approach. Specifically, we design a structured template prompt (Figure~\ref{prompt:template}) consisting of four parts concatenated in a fixed order (Figure~\ref{prompt:template-role-and-task},~\ref{prompt:template-transcript-format},~\ref{prompt:template-guidelines-format},~\ref{prompt:template-response-format}, and~\ref{prompt:template-instructions}). For the first part, we follow prior work showing that explicit role assignment can improve LLM performance~\cite{hu2024quantifying}. The remaining parts are designed to enforce a structured representation of inputs, constrain the output format, and provide unambiguous instructions. These parts are then paired with both the full transcript of an interview and the corresponding MADRS guideline section under examination.

In practice, given the transcript of an interview, a LLM is prompted independently for each of the 10 SIGMA sections. The LLM predicts the first and last utterance indices corresponding to each section. We then reconstruct the final annotation by assigning all utterances between the predicted \textit{start} and \textit{end} indices to the respective SIGMA section, resulting in the final segmentation of the transcript.

\begin{tcolorbox}[
  title=\textcolor{white}{PROMPT TEMPLATE},
  colback=blue!2,
  colframe=blue!60!black,
  boxrule=1pt,
  arc=2mm,
  sharp corners=south,
  fonttitle=\bfseries,
  coltitle=black,
  enhanced,
  breakable
]
\begin{verbatim}
{Role and task}

TRANSCRIPTION FORMAT:
{Transcript format}

GUIDELINES FORMAT:
{Guidelines format}

RESPONSE FORMAT:
{Response format}

INSTRUCTIONS:
{Instructions}
---------------------
INPUT TRANSCRIPT:
{transcript}

INPUT GUIDELINES:
{sigma section}

RESPONSE:
\end{verbatim}
\end{tcolorbox}
\captionof{figure}{Prompt template used for \textit{transcript segmentation}.}
\label{prompt:template}

\begin{tcolorbox}[title=Role and Task:]
\small
You are a blinded clinical expert independently reviewing clinical assessment interviews.
You will be provided with:
\begin{itemize}\vspace{-0.2cm}
\item[-] a transcript of an audio-recorded interview in which a 'clinician' interviews a 'patient'.\vspace{-0.2cm}
\item[-] a specific guidelines section that should be covered during the interview.
\end{itemize}\vspace{-0.2cm}
Your task is to identify and extract the segments of the transcript that correspond to a specified section of the clinical interview guidelines. 
\end{tcolorbox}
\captionof{figure}{Prompt section used for defining \textit{role and task}.}
\label{prompt:template-role-and-task}

\begin{tcolorbox}[title=Transcript Format:]
\small
TRANSCRIPT FORMAT:
\begin{verbatim}
[
  {"speaker": "clinician" | "patient",
   "transcription": "<text>",
   "id": <integer>
   },
  ...
]
\end{verbatim}
\end{tcolorbox}
\captionof{figure}{Prompt section used for defining \textit{transcript format}.}
\label{prompt:template-transcript-format}

\begin{tcolorbox}[title=Guidelines Format:]
\small
GUIDELINES FORMAT:
\begin{verbatim}
{
    "label": "<title of the section>",
    "script": [
        "<key questions that the 
          clinician should ask the
          patient>"
    ],
    "probes": [
        "<follow-up questions that
        may be used when further 
        exploration or additional 
        clarification of symptoms 
        is necessary>"
    ]
}
\end{verbatim}
\end{tcolorbox}
\captionof{figure}{Prompt section used for defining \textit{guidelines format}.}
\label{prompt:template-guidelines-format}

\begin{tcolorbox}[title=Response Format:]
\small
RESPONSE FORMAT: 

If one or more transcript segments are found: \begin{verbatim}[{"id_start": X, "id_end": Y}, ...]\end{verbatim}
If not: None

You must respond with a valid JSON object only. Do not include any additional text.
\end{tcolorbox}
\captionof{figure}{Prompt section used for defining \textit{response format}.}
\label{prompt:template-response-format}

\begin{tcolorbox}[title=Instructions:]
\small
INSTRUCTION:
\begin{itemize}\vspace{-0.2cm}
\item[-] Review the transcript and identify all segments that directly address the provided guidelines section (including both script and probe questions).\vspace{-0.2cm}
\item[-] For each continuous segment covering the guidelines section, return the corresponding first and last transcript record IDs as "id\_start" and "id\_end".\vspace{-0.2cm}
\item[-] Only include transcript segments that directly address the guidelines section. Do not include content that is merely thematically similar or unrelated.\vspace{-0.2cm}
\item[-] If multiple, non-overlapping segments address the guidelines section, return each as a separate {"id\_start", "id\_end"} pair.\vspace{-0.2cm}
\item[-] Be aware that:\vspace{-0.2cm}
\begin{itemize}
    \item[-] Speaker attribution may occasionally be incorrect.
    \item[-] Questions and answers may be split across multiple records, and unrelated or fragmented text may appear between them. In such cases, extend the segment range forward as needed to include the corresponding answer to a relevant question.
\end{itemize}\vspace{-0.2cm}
\item[-] If the guidelines section is not covered in the transcript, return no segments. \vspace{-0.2cm}
\end{itemize}

Review the transcript and identify the records that correspond to the provided guidelines section.  
For each continuous coverage of the guidelines section, return the first and last transcript records IDs (based on their id) that address the guidelines section.
\end{tcolorbox}
\captionof{figure}{Prompt section used for defining \textit{intructions}.}
\label{prompt:template-instructions}

\section{MADRS assessment}
\paragraph{LLAMADRS}
To compare with the state of the art, we evaluate decoder LLMs using the zero-shot LLAMADRS prompt proposed by~\citep{kebe2026llamadrs}. For fair comparisons, we evaluate LLAMADRS on the same Test set used for encoders. Figure~\ref{prompt:llamadrs} reports the prompt used in our experiments, as originally defined in the referenced study.

\begin{tcolorbox}[
  title=\textcolor{white}{LLAMADRS TEMPLATE},
  colback=blue!2,
  colframe=blue!60!black,
  boxrule=1pt,
  arc=2mm,
  sharp corners=south,
  fonttitle=\bfseries,
  coltitle=black,
  enhanced,
  breakable
]
\small
\begin{verbatim}
Task: Analyze a diarized transcript of a 
psychiatric session where the MADRS is 
being administered. Predict the rating (0-6) 
that the practitioner would likely give for 
the specified MADRS item based on the 
patient's responses and the conversation. 
Remember that the rating is for the last week,
not based on the patient's history or general 
condition. Focus on the questions that are 
directly related to the MADRS item and the 
patient's answers.

MADRS Item: {MADRS-ITEM}
  
[Descriptive Cues]
Description: Representing reports of depressed 
mood, regardless of whether it is reflected in 
appearance or not. Includes low spirits, 
despondency or the feeling of being beyond help 
and without hope. Rate according to intensity, 
duration and the extent to which the mood is 
reported to be influenced by events.
  
Rating Scale:
0 - Occasional sadness in keeping with the 
    circumstances.
1 - Between 0 and 2.
2 - Sad or low but brightens up without
    difficulty.
3 - Between 2 and 4.
4 - Pervasive feelings of sadness or gloominess. 
    The mood is still influenced by external
   circumstances.
5 - Between 4 and 6.
6 - Continuous or unvarying sadness, misery or 
    despondency.
  
Instructions:
1. Read the provided interview transcript.
2. Rate the interviewee’s reported sadness on a 
   scale of 0-6.
3. Provide a brief explanation (2-3 sentences) 
   for your rating.
4. List 2-3 key utterances from the conversation 
   supporting your assessment.
5. Return only a valid json.
  
Output Schema: 
{rating: 0-6, explanation: 2-3 sentences, 
key_utterances: [line numbers], 
most_relevant_question:
[...]}
\end{verbatim}
\end{tcolorbox}
\captionof{figure}{LLAMADRS prompt template used for \textit{MADRS assessment}.}
\label{prompt:llamadrs}

\paragraph{Encoder fine-tuning}\label{app:encoder-fine-tuning}
All encoders are fine-tuned on SIGMA segments using a Train/Dev/Test split, with no patient appearing in more than one set. For all experiments, the input text includes both clinician and patient turns from the extracted SIGMA segment. Specifically, we use this format: \texttt{\footnotesize clinician: [utterance] patient: [utterance]}. Each SIGMA segment is paired with the MADRS \textit{reviewer} score for the assessed item in the same interview. Item scores are modeled on the standard MADRS ordinal scale $y \in \{0,\ldots,6\}$.

\ \ \ For setting \textbf{(i)}, we use a shared transformer encoder with item-specific prediction heads. Given a segment $s_i$, the encoder produces a representation which is passed to the prediction head corresponding to MADRS item $m_i$. The model outputs six ordinal-threshold logits $z_{k}$, where $k \in \{0,\ldots,5\}$. Each MADRS score is represented using six binary threshold targets: 
\[
t_{i,k} = \mathbf{1}\{y_i > k\} \ .
\]
The logits are converted into threshold probabilities using the sigmoid activation function:
\[
p_{i,k} = \sigma(z_{i,k}) \ .
\]

Let $B$ denote the batch size, and let $w_{i,k}$ denote the positive-class weight applied to threshold $k$ for MADRS item $m_i$, used to account for class imbalance among the ordinal threshold targets.
The ordinal objective is binary cross-entropy over the six thresholds:
\[
\begin{aligned}
\mathcal{L}_{o}
= -\frac{1}{B}\sum_{n=1}^{B}\frac{1}{6}\sum_{k=0}^{5}
\Big[
& w_{i_n,k}t_{n,k}\log p_{n,k} \ + \\
& (1-t_{n,k})\log(1-p_{n,k})
\Big] \ .
\end{aligned}
\]
where $i_n$ denotes the MADRS item associated with sample $n$. At inference time, the predicted score is obtained by summing the thresholds whose predicted probability is at least 0.5:
\[
\hat{y}_i = \sum_{k=0}^{5}\mathbf{1}\{p_{i,k} \ge 0.5\}.
\]

\ \ \ For setting \textbf{(ii)}, independent models are fine-tuned with the same ordinal objective separately for each MADRS item. For setting \textbf{(iii)}, we use the architecture of \textbf{(i)} but replace the ordinal objective with a seven-class cross-entropy loss:
\[
\begin{aligned}
\mathcal{L}_{\mathrm{ce}}
= -\frac{1}{B}\sum_{n=1}^{B}
\alpha_{i_n,y_n}
\log
\frac{\exp z_{n,y_n}}
{\sum_{c=0}^{6}\exp z_{n,c}},
\end{aligned}
\]
where $\alpha_{i_n,y_n}$ is the item-specific class weight.

\ \ \ For \texttt{RoBERTa}, \texttt{MentalBERT}, \texttt{PubMedBERT}, and \texttt{ClinicalBERT}, we use a maximum input length of 512 tokens. Longer item segments are processed with sliding windows using a stride of 128 tokens. If an item segment yields $M_n$ windows, the row representation is the mean of the first-token window embeddings:
\[
h_n = \frac{1}{M_n}\sum_{m=1}^{M_n} H_{\theta}(x_{n,m})_0.
\]
The resulting segment-level vector is then passed to the prediction head for the corresponding MADRS item. \texttt{ModernBERT} is fine-tuned with a 4096-token context window, so each segment can be processed without sliding-window aggregation. For each window or full segment, we use the hidden state of the first special token as the sequence representation: \texttt{[CLS]} for BERT-style tokenizers and \texttt{<s>} for RoBERTa-style tokenizers. We apply dropout with rate 0.1 to this representation before the item-specific prediction head.

\begin{table}[t]
\centering
\small
\setlength{\tabcolsep}{4pt}
\renewcommand{\arraystretch}{1.08}
\begin{tabularx}{\columnwidth}{@{}l>{\raggedleft\arraybackslash}X@{}}
\toprule
\textbf{Hyperparameter} & \textbf{Value} \\
\midrule
Optimizer & AdamW \\
Learning rate & $1\times10^{-5}$ \\
Weight decay & 0.01 \\
Warmup ratio & 0.10 \\
Maximum epochs & 10 \\
Early stopping patience & 2 validation epochs \\
Gradient clipping & 1.0 \\
Dropout & 0.1 \\
Precision & FP32 \\
Random seed & 13 \\
Checkpoint selection & Lowest validation item MAE \\
Training batch size & 2 \ (1 for \texttt{ModernBERT}) \\
Gradient accumulation steps & 2 \ (4 for \texttt{ModernBERT}) \\
Evaluation batch size & 4 \ (2 for \texttt{ModernBERT}) \\
Effective training batch size & 4 \\
\bottomrule
\end{tabularx}
\caption{Fine-tuning hyperparameters.}
\label{tab:encoder-finetuning-hparams}
\end{table}
 
\section{Quality assessment}\label{app:quality-assessment}
\paragraph{Encoder-generated \textit{reference} scores}
Directly using the encoder models trained for \textit{MADRS scoring} would introduce information leakage: the MADRS ground-truth annotations of an interview used during \texttt{RoBERTa} training would subsequently influence the generation of \textit{reference} scores for the same interview during \textit{quality assessment}. 

\ \ \ To prevent this issue, we repeat the fine-tuning excluding the interviews from the \textit{quality assessment} subset. The resulting model is then used to generate MADRS predictions for those specific interviews. 

\ \ \ This procedure ensures that every generated MADRS score used for \textit{quality assessment} is produced by a model that has not observed the corresponding interview during training. Thus, the generated \textit{reference} scores represent out-of-sample predictions rather than reconstructions influenced by exposure to the target interviews.

\subsection{Error tolerance estimation}
When the \textit{reference} scores $\{r_{1},\ldots,r_{10}\}$ are generated by \madrsnospace, we account for the possibility that they may themselves contain item-level prediction errors. Consequently, we do not interpret every disagreement between a \textit{rater} and the generated scores as evidence of poor rater quality. Instead, we estimate an item-specific error tolerance using the Train set.

\ \ \ Let $\{r_{1,g},\ldots,r_{10,g}\}$, $\{r_{1,r},\ldots,r_{10,r}\}$, and $\{\hat{r}_{1},\ldots,\hat{r}_{10}\}$ denote the MADRS item scores generated by \madrsnospace, assigned by the \textit{reviewer} team, and assigned by the clinical \textit{rater}, respectively.

\ \ \ For each MADRS item $i$ and instance $j$ in the Train set, we compute the empirical generated-score error as the absolute difference between the generated score $r_{i,g}^{(j)}$ and the corresponding \textit{reviewer} score $r_{i,r}^{(j)}$:

\[
e_i^{(j)}=\left|r_{i,g}^{(j)}-r_{i,r}^{(j)}\right|.
\]

\ \ \ The item-specific error tolerance for MADRS item $i$ is defined as the noise threshold $\tau_i$, computed as the 80th percentile of the empirical generated-score errors for item $i$ across all Train-set instances:

\[
\tau_i=Q_{0.8}\left(\left\{e_i^{(j)}\right\}_{j=1}^{N_{\mathrm{train}}}\right),
\]

where $N_{\mathrm{train}}$ denotes the number of instances in the Train set. This threshold represents the expected prediction variability of the generated reference score for MADRS item $i$.

\ \ \ For our Random Forest classifiers, we define the excess-gap feature for item $i$ as the absolute score difference between \madrs and the clinical \textit{rater} that exceeds the expected noise level:

\[
d_i=\max\left(\left|r_{i,g}-\hat{r}_i\right|-\tau_i,0\right),
\]

where $r_{i,g}$ is the MADRS score generated by \madrsnospace, $\hat{r}_i$ is the score assigned by the clinical \textit{rater}, and $\tau_i$ is the item-specific noise threshold estimated from the Train set. Thus, score differences within the expected error range of the generated reference score are removed, while only the excess component beyond this tolerance is retained.

\ \ \ This procedure prevents the classifier from attributing expected uncertainty in the generated reference scores to the clinical \textit{rater}. The classifier is then trained using the item-level excess-gap features ${d_i}_{i=1}^{10}$ together with summary statistics, including the total excess gap and the mean excess gap across MADRS items:

\[
D_{\mathrm{sum}}=\sum_{i=1}^{10}d_i,
\]

\[
D_{\mathrm{mean}}=\frac{1}{10}\sum_{i=1}^{10}d_i.
\]

\subsection{Random forest parameters}
We use the default parameters provided by \texttt{scikit-learn} for the Random Forest classifier, varying only the number of estimators, which is set to 500.

\section{Evaluation results}\label{app:additional-results}
For comparison, we report in Table~\ref{tab:app_spearman}, Table~\ref{tab:app_mae}, and Table~\ref{tab:app_acc} the performance of \textit{MADRS assessment} for each MADRS item and the final MADRS total score, using Spearman correlation, MAE, and accuracy metrics, respectively. 
%For baselines, results are averaged over three runs.
%LLAMADRS~\cite{kebe2026llamadrs}
%\ref{tab:app_spearman}, \ref{tab:app_mae}, \ref{tab:app_acc}

% ===================== Spearman =====================
\begin{table*}[!ht]
\centering
\resizebox{\textwidth}{!}{%
\begin{tabular}{c|ccc|ccccc}
\multicolumn{1}{c|}{} &
\multicolumn{3}{c|}{\textbf{Baselines:} LLAMADRS~\cite{kebe2026llamadrs}} &
\multicolumn{5}{c}{\textbf{Encoders}} \\

\textbf{MADRS Items} &
\texttt{GPT-4.1} &
\texttt{Claude Sonnet 4.5} &
\texttt{Llama 3.3 70B} &
\texttt{RoBERTa} &
\texttt{ModernBERT} &
\texttt{MentalBERT} &
\texttt{ClinicalBERT} &
\texttt{PubMedBERT} \\ \hline

APPARENT SADNESS & 0.644 & 0.641 & 0.648 & \textbf{0.660} & 0.534 & 0.650 & 0.575 & 0.522 \\
REPORTED SADNESS & 0.650 & \textbf{0.657} & 0.649 & 0.632 & 0.485 & 0.622 & 0.393 & 0.378 \\
INNER TENSION & 0.762 & \textbf{0.767} & 0.760 & 0.698 & 0.668 & 0.713 & 0.651 & 0.664 \\
REDUCED SLEEP & 0.788 & \textbf{0.795} & 0.789 & 0.624 & 0.453 & 0.608 & 0.485 & 0.501 \\
REDUCED APPETITE & \textbf{0.832} & 0.828 & 0.827 & 0.752 & 0.599 & 0.700 & 0.671 & 0.684 \\
CONCENTRATION DIFFICULTIES & 0.717 & 0.715 & \textbf{0.721} & 0.686 & 0.691 & 0.692 & 0.680 & 0.687 \\
LASSITUDE & \textbf{0.679} & 0.673 & 0.674 & 0.675 & 0.528 & 0.622 & 0.518 & 0.541 \\
INABILITY TO FEEL & \textbf{0.681} & 0.680 & \textbf{0.681} & 0.611 & 0.478 & 0.590 & 0.499 & 0.494 \\
PESSIMISTIC THOUGHTS & 0.709 & 0.713 & 0.712 & 0.694 & 0.655 & \textbf{0.716} & 0.666 & 0.688 \\
SUICIDAL THOUGHTS & 0.588 & 0.586 & 0.567 & 0.812 & 0.820 & 0.813 & 0.814 & \textbf{0.841} \\ \hline

TOTAL\_SCORE & 0.849 & 0.848 & 0.851 & \textbf{0.867} & 0.716 & 0.842 & 0.754 & 0.756 \\ \hline

\end{tabular}}
\caption{\textbf{Spearman correlation}: performance comparison for \textit{MADRS assessment}.}
\label{tab:app_spearman}
\end{table*}

% ===================== MAE =====================
\begin{table*}[!ht]
\centering
\resizebox{\textwidth}{!}{%
\begin{tabular}{c|ccc|ccccc}
\multicolumn{1}{c|}{} &
\multicolumn{3}{c|}{\textbf{Baselines:} LLAMADRS~\cite{kebe2026llamadrs}} &
\multicolumn{5}{c}{\textbf{Encoders}} \\

\textbf{MADRS Items} &
\texttt{GPT-4.1} &
\texttt{Claude Sonnet 4.5} &
\texttt{Llama 3.3 70B} &
\texttt{RoBERTa} &
\texttt{ModernBERT} &
\texttt{MentalBERT} &
\texttt{ClinicalBERT} &
\texttt{PubMedBERT} \\ \hline

APPARENT SADNESS & 0.878 & 0.888 & 0.889 & \textbf{0.562} & 0.700 & 0.645 & 0.660 & 0.665 \\
REPORTED SADNESS & 0.676 & 0.670 & 0.683 & \textbf{0.483} & 0.621 & 0.557 & 0.773 & 0.803 \\
INNER TENSION & 0.805 & 0.795 & 0.807 & 0.527 & 0.695 & \textbf{0.512} & 0.714 & 0.685 \\
REDUCED SLEEP & 0.829 & 0.811 & 0.833 & \textbf{0.778} & 1.197 & 0.862 & 1.113 & 1.089 \\
REDUCED APPETITE & 0.748 & 0.759 & 0.756 & \textbf{0.709} & 0.901 & 0.793 & 0.862 & 0.857 \\
CONCENTRATION DIFFICULTIES & 0.600 & 0.594 & 0.602 & 0.564 & 0.559 & \textbf{0.554} & 0.574 & 0.578 \\
LASSITUDE & 0.722 & 0.733 & 0.733 & \textbf{0.623} & 0.824 & 0.667 & 0.853 & 0.848 \\
INABILITY TO FEEL & 0.885 & 0.889 & 0.894 & \textbf{0.725} & 0.819 & 0.755 & 0.868 & 0.868 \\
PESSIMISTIC THOUGHTS & 0.860 & 0.852 & 0.859 & 0.659 & 0.717 & \textbf{0.639} & 0.766 & 0.678 \\
SUICIDAL THOUGHTS & 0.907 & 0.915 & 0.964 & 0.322 & 0.341 & 0.322 & 0.322 & \textbf{0.312} \\ \hline

TOTAL\_SCORE & 4.044 & 4.080 & 4.133 & \textbf{2.956} & 4.200 & 3.288 & 4.322 & 4.161 \\ \hline

\end{tabular}}
\caption{\textbf{Mean Absolute Error}: performance comparison for \textit{MADRS assessment}. Item rows report item-level MAE; the final row reports MAE over summed total MADRS scores.}
\label{tab:app_mae}
\end{table*}

% ===================== Accuracy@1 =====================
\begin{table*}[!ht]
\centering
\resizebox{\textwidth}{!}{%
\begin{tabular}{cccc|ccccc}

\multicolumn{1}{c}{} &
\multicolumn{3}{c|}{\textbf{Baselines:} LLAMADRS~\cite{kebe2026llamadrs}} &
\multicolumn{5}{c}{\textbf{Encoders}} \\

\textbf{MADRS Items} &
\texttt{GPT-4.1} &
\texttt{Claude Sonnet 4.5} &
\texttt{Llama 3.3 70B} &
\texttt{RoBERTa} &
\texttt{ModernBERT} &
\texttt{MentalBERT} &
\texttt{ClinicalBERT} &
\texttt{PubMedBERT} \\ \hline

APPARENT SADNESS & 0.794 & 0.790 & 0.789 & \textbf{0.906} & 0.872 & 0.887 & 0.892 & 0.892 \\
REPORTED SADNESS & 0.855 & 0.859 & 0.855 & 0.946 & 0.906 & \textbf{0.916} & 0.857 & 0.857 \\
INNER TENSION & 0.828 & 0.829 & 0.826 & 0.941 & 0.872 & \textbf{0.946} & 0.862 & 0.872 \\
REDUCED SLEEP & 0.807 & \textbf{0.811} & 0.810 & 0.803 & 0.729 & 0.793 & 0.724 & 0.729 \\
REDUCED APPETITE & 0.842 & 0.837 & 0.837 & \textbf{0.867} & 0.754 & 0.823 & 0.764 & 0.754 \\
CONCENTRATION DIFFICULTIES & 0.857 & 0.863 & 0.859 & \textbf{0.892} & 0.868 & 0.873 & 0.853 & 0.858 \\
LASSITUDE & 0.852 & 0.847 & 0.847 & \textbf{0.902} & 0.824 & 0.897 & 0.814 & 0.799 \\
INABILITY TO FEEL & 0.785 & 0.784 & 0.782 & \textbf{0.882} & 0.828 & 0.828 & 0.804 & 0.814 \\
PESSIMISTIC THOUGHTS & 0.745 & 0.750 & 0.748 & \textbf{0.873} & 0.834 & 0.883 & 0.810 & 0.839 \\
SUICIDAL THOUGHTS & 0.766 & 0.764 & 0.753 & \textbf{0.941} & 0.932 & \textbf{0.941} & 0.937 & \textbf{0.941} \\ \hline

TOTAL\_SCORE & 0.813 & 0.814 & 0.811 & \textbf{0.895} & 0.879 & 0.842 & 0.836 & 0.832 \\ \hline
\end{tabular}}
\caption{\textbf{Accuracy@1}: performance comparison for \textit{MADRS assessment}. Item rows report item-level within-one accuracy; the final row reports Acc@1 over summed total MADRS scores.}
\label{tab:app_acc}
\end{table*}

\section{Human clinical assessment}\label{app:ira}
All assessments were obtained through real-time, interviews conducted by trained clinicians in private clinics outside the scope of this work. Following standard practices in clinical trials, interviews were subsequently reviewed and, where necessary, adjudicated by a senior clinician based on audio recordings of the original interviews~\cite{lipsitz2004rater}. Inter-rater agreement between interviewing and audio-reviewing clinicians is reported in Table~\ref{tab:ira}, computed using Spearman’s correlation~\cite{spearman1904general}, Quadratic Weighted Kappa (QWK)~\cite{cohen1968weighted}, and Intraclass Correlation Coefficient (ICC)~\cite{shrout1979intraclass}. Although the ratings are not strictly independent (the audio-reviewing clinician had access to the original interview recordings), high values across all measures indicate strong concordance between the original and adjudicated scores. A subset of these recordings was re-analyzed for quality assessment, where the difference between the initial \textit{rater} and \textit{reviewer} scores was used to define a binary quality label.

\begin{table}[h]
\resizebox{\columnwidth}{!}{%
\begin{tabular}{lccc}
\hline
\textbf{MADRS Items} & \multicolumn{3}{c}{\textbf{DATASET}} \\
 & QWK & ICC & Spearm. \\
\hline
APPARENT SADNESS & 0.940 & 0.937 & 0.898 \\
REPORTED SADNESS & 0.944 & 0.944 & 0.912 \\
INNER TENSION & 0.946 & 0.946 & 0.919 \\
REDUCED SLEEP & 0.950 & 0.944 & 0.929 \\
REDUCED APPETITE & 0.958 & 0.958 & 0.947 \\
CONC. DIFFICULTIES & 0.955 & 0.955 & 0.933 \\
LASSITUDE & 0.943 & 0.943 & 0.914 \\
INABILITY TO FEEL & 0.918 & 0.914 & 0.866 \\
PESSIMISTIC THOUGHTS & 0.937 & 0.937 & 0.933 \\
SUICIDAL THOUGHTS & 0.895 & 0.895 & 0.904 \\
\hline
\textbf{TOTAL\_SCORE} & 0.980 & 0.980 & 0.963 \\
\hline
\end{tabular}}
\caption{Inter-rater agreement between the interviewing clinician and the audio-reviewing clinician across all studies.}
\label{tab:ira}
\end{table}

\end{document}